\documentclass[runningheads]{llncs}
\usepackage{graphicx}
\usepackage{amsmath,amssymb} 
\usepackage{booktabs}
\usepackage{url}
\usepackage{pifont}

\newcommand{\cmark}{\ding{51}}%
\newcommand{\xmark}{\ding{55}}%
\usepackage{algorithm}
\usepackage{algpseudocode}

\makeatletter
\def\BState{\State\hskip-\ALG@thistlm}
\makeatother

\usepackage{color}
\usepackage{xcolor}
\usepackage{colortbl}

\newcommand{\tabincell}[2]{\begin{tabular}{@{}#1@{}}#2\end{tabular}}  
\definecolor{Gray}{gray}{0.9}
\newcolumntype{g}{>{\columncolor{Gray}}l}

\usepackage[symbol]{footmisc}

\newcommand*\samethanks[1][\value{footnote}]{\footnotemark[#1]}

\begin{document}


\pagestyle{headings}
\mainmatter

\title{Simple Baselines for Human Pose Estimation and Tracking} 

\titlerunning{Simple Baselines for Human Pose Estimation and Tracking}

\author{Bin Xiao\inst{1}\thanks{Equal contribution.} \and
Haiping Wu\inst{2}\samethanks[1]\thanks{This work is done when Haiping Wu is an intern at Microsoft Research Asia.} \and
Yichen Wei\inst{1}}
%
\authorrunning{B. Xiao et al.}
%

\institute{Microsoft Research Asia, \and
University of Electronic Science and Technology of China \\
\email{\{Bin.Xiao, v-haipwu, yichenw\}@microsoft.com}}

\maketitle

\begin{abstract}
There has been significant progress on pose estimation and increasing interests on pose tracking in recent years. At the same time, the overall algorithm and system complexity increases as well, making the algorithm analysis and comparison more difficult. This work provides simple and effective baseline methods. They are helpful for inspiring and evaluating new ideas for the field. State-of-the-art results are achieved on challenging benchmarks. The code will be available at \url{https://github.com/leoxiaobin/pose.pytorch}.
\end{abstract}

\section{Introduction}
Similar as many vision tasks, the progress on human pose estimation problem is significantly advanced by deep learning. Since the pioneer work in~\cite{toshev2014deeppose,tompson2014joint}, the performance on the MPII benchmark~\cite{andriluka14cvpr} has become saturated in three years, starting from about $80\%$ PCKH$@$0.5~\cite{tompson2014joint} to more than $90\%$~\cite{newell2016stacked,chu2017multi,chen2017adversarial,yang2017learning}. The progress on the more recent and challenging COCO human pose benchmark~\cite{lin2014microsoft} is even faster. The mAP metric is increased from 60.5 (COCO 2016 Challenge winner~\cite{cocoleaderboard,cao2017realtime}) to 72.1(COCO 2017 Challenge winner~\cite{chen2018cascaded,cocoleaderboard}) in one year. With the quick maturity of pose estimation, a more challenging task of ``simultaneous pose detection and tracking in the wild'' has been introduced recently~\cite{andriluka2018posetrack}.

At the same time, the network architecture and experiment practice have steadily become more complex. This makes the algorithm analysis and comparison more difficult. For example, the leading methods~\cite{newell2016stacked,chu2017multi,chen2017adversarial,yang2017learning} on MPII benchmark~\cite{andriluka14cvpr} have considerable difference in many details but minor difference in accuracy. It is hard to tell which details are crucial. Also, the representative works~\cite{newell2017associative,papandreou2017towards,he2017mask,chen2018cascaded,cao2017realtime} on COCO benchmark are also complex but differ significantly. Comparison between such works is mostly on system level and less informative. About pose tracking, although there has not been much work~\cite{andriluka2018posetrack}, the system complexity can be expected to further increase due to the increased problem dimension and solution space.

This work aims to ease this problem by asking a question from the opposite direction, \emph{how good could a simple method be?} To answer the question, this work provides baseline methods for both pose estimation and tracking. They are quite simple but surprisingly effective. Thus, they hopefully would help inspiring new ideas and simplifying their evaluation. The code, as well as pre-trained models, will be released to facilitate the research community.

Our pose estimation is based on a few deconvolutional layers added on a backbone network, ResNet~\cite{he2016deep} in this work. It is probably the simplest way to estimate heat maps from deep and low resolution feature maps. Our \emph{single} model's best result achieves the state-of-the-art at mAP of 73.7 on COCO test-dev split, which has an improvement of $1.6\%$ and $0.7\%$ over the winner of COCO 2017 keypoint Challenge's single model and their ensembled model~\cite{chen2018cascaded,cocoleaderboard}.

Our pose tracking follows a similar pipeline of the winner~\cite{girdhar2018detect} of ICCV'17 PoseTrack Challenge~\cite{andriluka2018posetrack}. The single person pose estimation uses our own method as above. The pose tracking uses the same greedy matching method as in~\cite{girdhar2018detect}. \emph{Our only modification is to use optical flow based pose propagation and similarity measurement}. Our best result achieves a mAP score of 74.6 and a MOTA score of 57.8, an absolute $15\%$ and $6\%$ improvement over 59.6 and 51.8 of the winner of ICCV'17 PoseTrack Challenge~\cite{girdhar2018detect,posetrackleaderboard}. It is the new state-of-the-art.

This work is not based on any theoretic evidence. It is based on simple techniques and validated by comprehensive ablation experiments, at our best. Note that we do not claim any algorithmic superiority over previous methods, in spite of better results. We do not perform complete and fair comparison with previous methods, because this is difficult and not our intent. As stated, the contribution of this work are solid baselines for the field.

\section{Pose Estimation Using A Deconvolution Head Network}
\label{sec:deconv_head}

\begin{figure}[t]
\centering
\includegraphics[width=12cm]{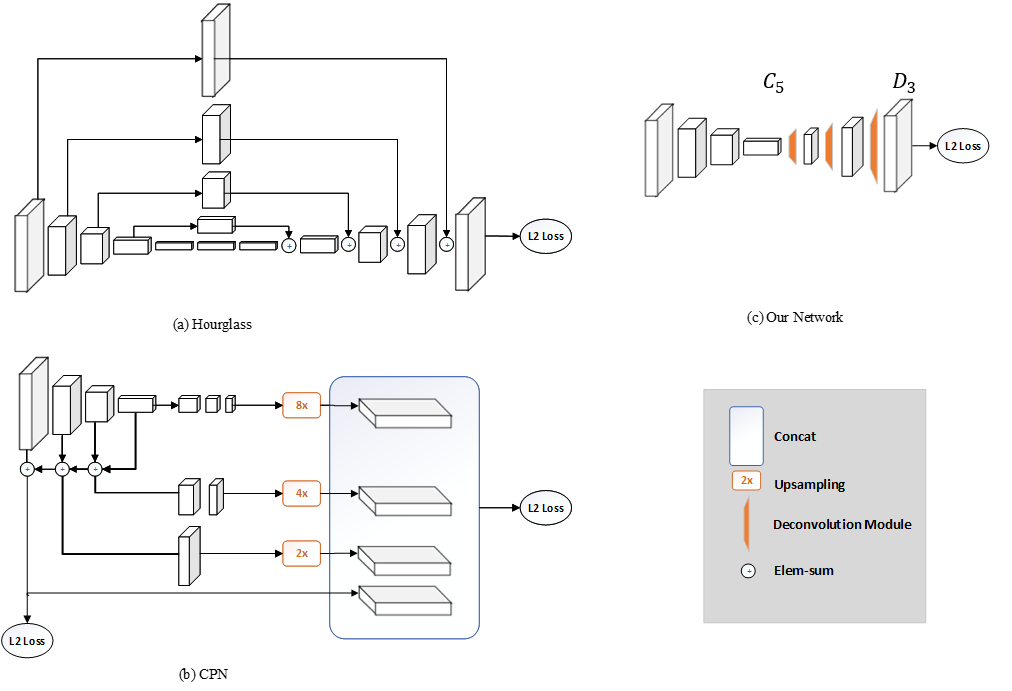}
\caption{Illustration of two state-of-the-art network architectures for pose estimation (a) one stage in Hourglass~\cite{newell2016stacked}, (b) CPN~\cite{chen2018cascaded}, and our simple baseline (c).}
\label{fig:deconv_pose}
\end{figure}

ResNet~\cite{he2016deep} is the most common backbone network for image feature extraction. It is also used in~\cite{papandreou2017towards,chen2018cascaded} for pose estimation. Our method simply adds a few deconvolutional layers over the last convolution stage in the ResNet, called $C_5$. The whole network structure is illustrated in Fig.~\ref{fig:deconv_pose}(c). We adopt this structure because it is arguably the simplest to generate heatmaps from deep and low resolution features and also adopted in the state-of-the-art Mask R-CNN~\cite{he2017mask}.

By default, three deconvolutional layers with batch normalization~\cite{ioffe2015batch} and ReLU activation~\cite{krizhevsky2012imagenet} are used. Each layer has 256 filters with $4\times4$ kernel. The stride is $2$. A $1\times1$ convolutional layer is added at last to generate predicted heatmaps $\{H_{1}\dots H_{k}\}$ for all $k$ key points.

Same as in~\cite{tompson2014joint,newell2016stacked}, Mean Squared Error (MSE) is used as the loss between the predicted heatmaps and targeted heatmaps. The targeted heatmap $\hat{H_{k}}$ for joint $k$ is generated by applying a 2D gaussian centered on the $k^{th}$ joint's ground truth location.

\paragraph{Discussions} To understand the simplicity and rationality of our baseline, we discuss two state-of-the-art network architectures as references, namely, Hourglass~\cite{newell2016stacked} and CPN~\cite{chen2018cascaded}. They are illustrated in Fig.~\ref{fig:deconv_pose}.

Hourglass~\cite{newell2016stacked} is the dominant approach on MPII benchmark as it is the basis for all leading methods~\cite{chu2017multi,chen2017adversarial,yang2017learning}. It features in a multi-stage architecture with repeated bottom-up, top-down processing and skip layer feature concatenation.

Cascaded pyramid network (CPN)~\cite{chen2018cascaded} is the leading method on COCO 2017 keypoint challenge~\cite{cocoleaderboard}. It also involves skip layer feature concatenation and an online hard keypoint mining step.

Comparing the three architectures in Fig.~\ref{fig:deconv_pose}, it is clear that our method differs from~\cite{newell2016stacked,chen2018cascaded} in \emph{how high resolution feature maps are generated}. Both works~\cite{newell2016stacked,chen2018cascaded} use upsampling to increase the feature map resolution and put convolutional parameters in other blocks. In contrary, our method combines the upsampling and convolutional parameters into deconvolutional layers in a much simpler way, without using skip layer connections.

A commonality of the three methods is that three upsampling steps and also three levels of non-linearity (from the deepest feature) are used to obtain high-resolution feature maps and heatmaps. Based on above observations and the good performance of our baseline, it seems that \emph{obtaining high resolution feature maps is crucial, but no matter how}. Note that this discussion is only preliminary and heuristic. It is hard to conclude which architecture in Fig.~\ref{fig:deconv_pose} is better. This is not the intent of this work.

\begin{figure*}
	\centering
	\includegraphics[width=1.0\textwidth]{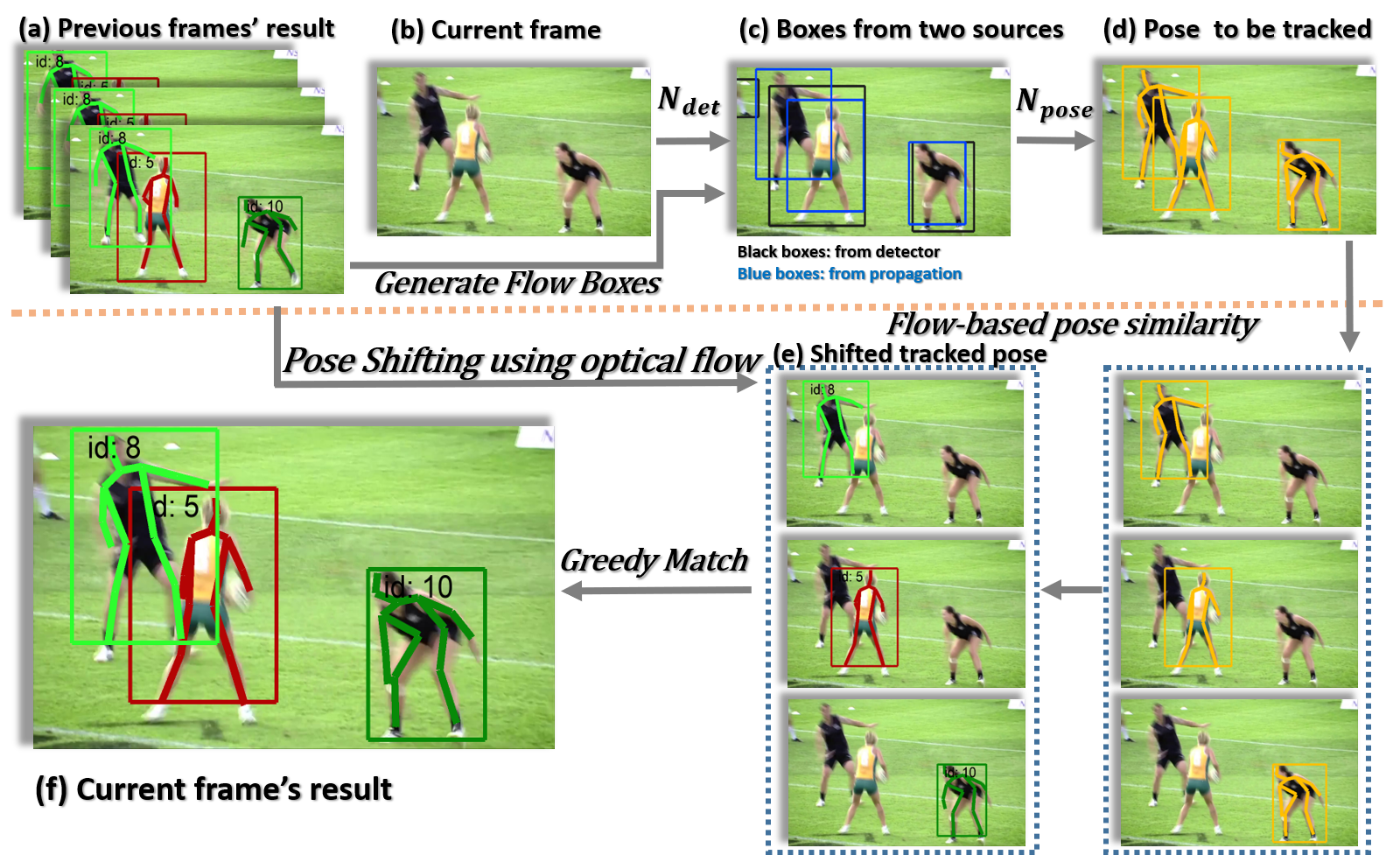}
	\caption{The proposed flow-based pose tracking framework.}
	\label{fig:tracking_framework}
\end{figure*}

\section{Pose Tracking Based on Optical Flow}
Multi-person pose tracking in videos first estimates human poses in frames, and then tracks these human pose by assigning a unique identification number ($id$) to them across frames. We present human instance $P$ with $id$ as $P=(J, id)$, where $J=\{j_{i}\}_{1:N_{J}}$ is the coordinates set of $N_{J}$ body joints and $id$ indicates the tracking id. When processing the $k^{th}$ frame $I^k$, we have the already processed human instances set $\mathcal{P}^{k-1} = \{P^{k-1}_{i}\}_{1:N_{k-1}}$ in frame $I^{k-1}$ and the instances set $\mathcal{P}^k = \{P^{k}_{i}\}_{1:N_{k}}$ in frame $I^k$ whose $id$ is to be assigned, where $N_{k-1}$ and $N_{k}$ are the instance number in frame $I^{k-1}$ and $I^{k}$. 
If one instance $P^{k}_{j}$ in current frame $I^{k}$ is linked to the instance $P^{k-1}_{i}$ in $I^{k-1}$ frame, then $id^{k-1}_{i}$ is propagated to $id^{k}_{j}$, otherwise a new $id$ is assigned to $P^{k}_{j}$, indicating a new track.

The winner~\cite{girdhar2018detect} of ICCV'17 PoseTrack Challenge~\cite{andriluka2018posetrack} solves this multi-person pose tracking problem by first estimating human pose in frames using Mask R-CNN~\cite{he2017mask}, and then performing online tracking using a greedy bipartite matching algorithm frame by frame. 

The greedy matching algorithm is to first assign the $id$ of $P^{k-1}_{i}$ in frame $I^{k-1}$ to $P^{k}_{j}$ in frame $I^k$ if the similarity between $P^{k-1}_{i}$ and $P^{k}_{j}$ is the highest, then remove these two instances from consideration, and repeat the $id$ assigning process with the highest similarity. When an instance $P^{k}_{j}$ in frame $I^k$ has no existing $P^{k-1}_{i}$ left to link, a new $id$ number is assigned, which indicates a new instance comes up.

We mainly follow this pipeline in~\cite{girdhar2018detect} with two differences. One is that we have 
two different kinds of human boxes, one is from a human detector and the other are boxes generated from previous frames using optical flow. The second difference is the similarity metric used by the greedy matching algorithm. We propose to use a flow-based pose similarity metric. 
Combined with these two modifications, we have our enhanced flow-based pose tracking algorithm, illustrated in Fig.~\ref{fig:tracking_framework}. We elaborate our flow-based pose tracking algorithm in the following.

\subsection{Joint Propagation using Optical Flow}
Simply applying a detector designed for single image level (e.g. Faster-RCNN~\cite{ren2015faster}, R-FCN~\cite{dai16rfcn})  to videos could lead to missing detections and false detections due to motion blur and occlusion introduced by video frames. As shown in Fig.~\ref{fig:tracking_framework}(c), the detector misses the left black person due to fast motion. Temporal information is often leveraged to generate more robust detections~\cite{zhu2017deep,zhu2017flow}.

We propose to generate boxes for the processing frame from nearby frames using temporal information expressed in optical flow.

Given one human instance with joints coordinates set $J^{k-1}_{i}$ in frame $I^{k-1}$ and the optical flow field $F_{k-1\rightarrow k}$ between frame $I^{k-1}$ and $I^k$, we could estimate the corresponding joints coordinates set $\hat{J^{k}_{i}}$ in frame $I^k$ by propagating the joints coordinates set $J^{k-1}_{i}$ according to $F_{k-1\rightarrow k}$. More specifically, for each joint location $(x,y)$ in $J^{k-1}_{i}$, the propagated joint location would be $(x+\delta{x}, y+\delta{y})$, where $\delta{x},\delta{y}$ are the flow field values at joint location $(x,y)$.
Then we compute a bounding of the propagated joints coordinates set $\hat{J^{k}_{i}}$, and expand that box by some extend (15\% in experiments) as the candidated box for pose estimation.
    
When the processing frame is difficult for human detectors that could lead to missing detections due to motion blur or occlusion, we could have boxes propagated from previous frames where people have been detected correctly. As shown in Fig.~\ref{fig:tracking_framework}(c), for the left black person in images, since we have the tracked result in previous frames in Fig.~\ref{fig:tracking_framework}(a), the propagated boxes successfully contain this person.

\subsection{Flow-based Pose Similarity} Using bounding box IoU(Intersection-over-Union) as the similarity metric ($S_{Bbox}$) to link instances could be problematic when an instance moves fast thus the boxes do not overlap, and in crowed scenes where boxes may not have the corresponding relationship with instances. A more fine-grained metric could be a pose similarity ($S_{Pose}$) which calculates the body joints distance between two instances using Object Keypoint Similarity (OKS). The pose similarity could also be problematic when the pose of the same person is different across frames due to pose changing. We propose to use a flow-based pose similarity metric.

Given one instance $J^{k}_{i}$ in frame $I^k$ and one instance $J^{l}_{j}$ in frame $I^l$, the flow-based pose similarity metric is represented as 
\begin{equation}
    S_{Flow}(J^{k}_{i}, J^{l}_{j}) = OKS(\hat{J^{l}_{i}}, J^{l}_{j}),
\end{equation}
where $OKS$ represents calculating the Object Keypoint Similarity (OKS) between two human pose, and $\hat{J^{l}_{i}}$ represents the propagated joints for $J^{k}_{i}$ from frame $I^k$ to $I^l$ using optical flow field $F_{k\rightarrow l}$. 
    
Due to occlusions with other people or objects, people often disappear and re-appear again. Considering consecutive two frames is not enough, thus we have the flow-based pose similarity considering multi frames, denoted as $S_{Multi-flow}$, meaning the propagated $\hat{J_k}$ comes from multi previous frames. In this way, we could relink instances even disappearing in middle frames.

\subsection{Flow-based Pose Tracking Algorithm}
With the joint propagation using optical flow and the flow-based pose similarity, we propose the flow-based pose tracking algorithm combining these two, as presented in Algorithm~\ref{alg.inference_unified}. Table~\ref{table:notations} summarizes the notations used in Algorithm~\ref{alg.inference_unified}.

\setlength{\tabcolsep}{2pt}
\begin{table}[t]
\begin{center}
\caption{Notations in Algorithm~\ref{alg.inference_unified}.}

\label{table:notations}
\begin{tabular}{l|l}
\hline
$I^{k}$&$k^{th}$ frame\\
$Q$&tracked instances queue\\
$L_{Q}$&max capacity of $Q$\\
$\mathcal{P}^{k}$&instances set in $k^{th}$ frame\\
$\mathcal{J}^{k}$&instances set of body joints in $k^{th}$ frame\\
$P^{k}_{i}$&$i^{th}$ instance in $k^{th}$ frame\\
$J^{k}_{i}$&body joints set of $i^{th}$ instance in $k^{th}$ frame\\
$F_{k\rightarrow l}$&flow field from $k^{th}$ frame to $l^{th}$ frame\\
$M_{sim}$&similariy matrix\\
\hline
$B^{k}_{\rm det}$&boxes from person detector in $k^{th}$ frame\\
$B^{k}_{\rm flow}$&boxes generated by joint propagating in $k^{th}$ frame\\
$B^{k}_{\rm unified}$&boxes unified by box $NMS$ in $k^{th}$ frame\\
\hline
$\mathcal{N}_{\rm det}$&person detection network\\
$\mathcal{N}_{\rm pose}$&human pose estimation network\\
$\mathcal{N}_{\rm flow}$&flow estimation network\\
\hline
$\mathcal{F}_{\rm sim}$&function for calculating similarity matrix\\
$\mathcal{F}_{\rm NMS}$&function for $NMS$ operation\\
$\mathcal{F}_{\rm FlowBoxGen}$&function for generating boxes by joint propagating\\
$\mathcal{F}_{\rm AssignID}$&function for assigning instance $id$\\
\hline
\end{tabular}
\end{center}
\end{table}
\setlength{\tabcolsep}{1.4pt}
    
\begin{algorithm}[h]  
\caption{The flow-based inference algorithm for video human pose tracking}  
\begin{algorithmic}[1]  
    \State \textbf{input}: video frames $\{I^k\}$, $Q=[]$, $Q$'s max capacity $L_{Q}$.
    \State $B^{0}_{\rm det} = \mathcal{N}_{\rm det}(I^0)$
    \State $\mathcal{J}^{0} = \mathcal{N}_{\rm pose}(I^0, B^{0}_{\rm det})$
    \State $\mathcal{P}^{0} = (\mathcal{J}^{0}, id)$ \Comment{initialize the $id$ for the first frame}
    \State $Q = [\mathcal{P}_0]$ \Comment{append the instance set $\mathcal{P}_{0}$ to $Q$}
    \For{$k=1$ \textbf{to} $\infty$}
    	\State $B^{k}_{\rm det} = \mathcal{N}_{\rm det}(I^{k})$ 
        \State $B^{k}_{\rm flow} = \mathcal{F}_{\rm FlowBoxGen}(\mathcal{J}^{k-1}, F_{k-1\rightarrow k})$ 
        \State $B^{k}_{\rm unified} = \mathcal{F}_{\rm NMS}(B^{k}_{\rm det}, B^{k}_{\rm flow})$ \Comment{unify detection boxes and flow boxes}
        \State $\mathcal{J}^{k} = \mathcal{N}_{\rm pose}(I^{k}, B^{k}_{\rm unified})$ 
        \State $M_{\rm sim} = \mathcal{F}_{\rm sim}(Q, \mathcal{J}^{k})$
        \State $\mathcal{P}^{k} = \mathcal{F}_{\rm AssignID}(M_{\rm sim}, \mathcal{J}^{k})$
        \State append $\mathcal{P}^{k}$ to $Q$ \Comment{update the $Q$}
    \EndFor
\end{algorithmic}  

\label{alg.inference_unified}  
\end{algorithm}  

First, we solve the pose estimation problem. For the processing frame in videos, the boxes from a human detector and boxes generated by propagating joints from previous frames using optical flow are unified using a bounding box Non-Maximum Suppression (NMS) operation. The boxes generated by progagating joints serve as the complement of missing detections of the detector (e.g. in Fig.~\ref{fig:tracking_framework}(c)). Then we estimate human pose using the cropped and resized images by these boxes through our proposed pose estimation network in Section~\ref{sec:deconv_head}.

Second, we solve the tracking problem. We store the tracked instances in a double-ended queue(Deque) with fixed length $L_{Q}$, denoted as 
\begin{equation}
Q = [ \mathcal{P}_{k-1},\mathcal{P}_{k-2},...,\mathcal{P}_{k-L_{Q}} ]
\end{equation}
where $\mathcal{P}_{k-i}$ means tracked instances set in previous frame $I^{k-i}$ and the $Q$'s length $L_{Q}$ indicates how many previous frames considered when performing matching. 

The $Q$ could be used to capture previous multi frames' linking relationship, initialized in the first frame in a video. For the $k^{th}$ frame $I^k$, we calculate the flow-based pose similarity matrix $M_{\rm sim}$
between the untracked instances set of body joints $\mathcal{J}^{k}$ (\emph{id} is none) and previous instances sets in $Q$ .
Then we assign \emph{id} to each body joints instance $J$ in $\mathcal{J}^{k}$ to get assigned instance set $\mathcal{P}^{k}$ by using greedy matching and $M_{\rm sim}$.
Finally we update our tracked instances $Q$ by adding up $k^{th}$ frame instances set $\mathcal{P}^{k}$.

\section{Experiments}
\subsection{Pose Estimation on COCO}
\label{sec:exp_coco}

The COCO Keypoint Challenge~\cite{lin2014microsoft} requires localization of multi-person keypoints in challenging uncontrolled conditions. The COCO train, validation, and test sets contain more than 200k images and 250k person instances labeled with keypoints. 150k instances of them are publicly available for training and validation. Our models are only trained on all COCO \emph{train2017} dataset (includes 57K images and 150K person instances) no extra data involved, ablation are studied on the \emph{val2017} set and finally we report the final results on \emph{test-dev2017} set to make a fair comparison with the public state-of-the-art results~\cite{cao2017realtime,he2017mask,papandreou2017towards,chen2018cascaded}. 

The COCO evaluation defines the object keypoint similarity (OKS) and uses the mean average precision (AP) over 10 OKS thresholds as main competition metric~\cite{cocoleaderboard}. The OKS plays the same role as the IoU in object detection. It is calculated from the distance between predicted points and ground truth points normalized by scale of the person.

\subsubsection{Training} The ground truth human box is made to a fixed aspect ratio, e.g., $height : width = 4 : 3$ by extending the box in height or width. It is then cropped from the image and resized to a fixed resolution. The default resolution is $256 : 192$. It is the same as the state-of-the-art method~\cite{chen2018cascaded} for a fair comparison. Data augmentation includes scale($\pm30\%$), rotation($\pm40$ degrees) and flip.

Our ResNet~\cite{he2016deep} backbone network is initialized by pre-training on ImageNet classification task~\cite{russakovsky2015imagenet}. In the training for pose estimation, the base learning rate is 1e-3. It drops to 1e-4 at 90 epochs and 1e-5 at 120 epochs. There are 140 epochs in total. Mini-batch size is 128. Adam~\cite{kingma2014adam} optimizer is used. Four GPUs on a GPU server is used. 

ResNet of depth 50, 101 and 152 layers are experimented. ResNet-50 is used by default, unless otherwise noted.

\subsubsection{Testing} A two-stage top-down paradigm is applied, similar as in~\cite{papandreou2017towards,chen2018cascaded}. For detection, by default we use a faster-RCNN~\cite{ren2015faster} detector with detection AP 56.4 for the person category on COCO \emph{val2017}. Following the common practice in~\cite{chen2018cascaded,newell2016stacked}, the joint location is predicted on the averaged heatmpaps of the original and flipped image. A quarter offset in the direction from highest response to the second highest response is used to obtain the final location. 

\setlength{\tabcolsep}{2pt}
\begin{table}[t]
\begin{center}
\caption{Ablation study of our method on COCO val2017 dataset. Those settings used in comparison are in \textbf{bold}. For example, (a, e, f) compares backbones.}

\label{table:coco_ablation_result}
\begin{tabular}{cllccc}
\hline\noalign{\smallskip}
Method & Backbone & Input Size &  $\#$Deconv. Layers & Deconv. Kernel Size& $AP$\\
\noalign{\smallskip}
\hline
$a$&\textbf{ResNet-50}&$\textbf{256}\times\textbf{192}$&\textbf{3}&\textbf{4}&70.4\\
$b$&ResNet-50&$256\times192$&\textbf{2}&4&67.9\\
\hline
$c$&ResNet-50&$256\times192$&3&\textbf{2}&70.1\\
$d$&ResNet-50&$256\times192$&3&\textbf{3}&70.3\\
\hline
$e$&\textbf{ResNet-101}&$256\times192$&3&4&71.4\\
$f$&\textbf{ResNet-152}&$256\times192$&3&4&72.0\\
\hline
$g$&ResNet-50&$\textbf{128}\times\textbf{96}$&3&4&60.6\\ 
$h$&ResNet-50&$\textbf{384}\times\textbf{288}$&3&4&72.2\\
\hline
\end{tabular}
\end{center}
\end{table}
\setlength{\tabcolsep}{1.4pt}

\subsubsection{Ablation Study}
Table~\ref{table:coco_ablation_result} investigates various options in our baseline in Section~\ref{sec:deconv_head}.

\begin{enumerate}
\item \emph{Heat map resolution.} Method (a) uses three deconvolutional layers to generate $64\times48$ heatmaps. Method (b) generates $32\times24$ heatmaps with two deconvolutional layers. (a) outperform (b) by 2.5 AP with only slightly increased model capacity. By default, three deconvolutional layers are used.

\item \emph{Kernel size.} Methods (a, c, d) show that a smaller kernel size gives a marginally decrease in AP, which is 0.3 point decrease from kernel size 4 to 2. By default, deconvolution kernel size of 4 is used.

\item \emph{Backbone.} As in most vision tasks, a deeper backbone model has better performance. Methods (a, e, f) show steady improvement by using deeper backbone models. AP increase is 1.0 from ResNet-50 to Resnet-101 and 1.6 from ResNet-50 to ResNet-152.

\item \emph{Image size.} Methods (a, g, h) show that image size is critical for performance. From method (a) to (g), the image size is reduced by half and AP drops points. On the other hand, relative $75\%$ computation is saved. Method (h) uses a large image size and increases 1.8 AP from method (a), at the cost of higher computational cost.
\end{enumerate}

\setlength{\tabcolsep}{4pt}
\begin{table}[t]
\begin{center}
\caption{Comparison with Hourglass~\cite{newell2016stacked} and CPN~\cite{chen2018cascaded} on COCO val2017 dataset. Their results are cited from~\cite{chen2018cascaded}. OHKM means Online Hard Keypoints Mining.}
\label{table:coco_ablation_com_with_other}
\begin{tabular}{lllccc}
\hline\noalign{\smallskip}
Method & Backbone & Input Size &  OHKM &  $AP$\\
\noalign{\smallskip}
\hline
8-stage Hourglass&-&$256\times192$&\xmark&66.9\\
8-stage Hourglass&-&$256\times256$&\xmark&67.1\\
CPN&ResNet-50&$256\times192$&\xmark&68.6\\
CPN&ResNet-50&$384\times288$&\xmark&70.6\\
CPN&ResNet-50&$256\times192$&\cmark&69.4\\
CPN&ResNet-50&$384\times288$&\cmark&71.6\\
\hline
Ours&ResNet-50&$256\times192$&\xmark&70.4\\ 
Ours&ResNet-50&$384\times288$&\xmark&72.2\\
\hline
\end{tabular}
\end{center}
\end{table}
\setlength{\tabcolsep}{1.4pt}

\subsubsection{Comparison with Other Methods on COCO val2017}

Table~\ref{table:coco_ablation_com_with_other} compares our results with a 8-stage Hourglass~\cite{newell2016stacked} and CPN~\cite{chen2018cascaded}. All the three methods use a similar top-down two-stage paradigm. For reference, the person detection AP of hourglass~\cite{newell2016stacked} and CPN~\cite{chen2018cascaded} is 55.3~\cite{chen2018cascaded}, which is comparable to ours 56.4. 

Compared with Hourglass~\cite{newell2016stacked,chen2018cascaded}, our baseline has an improvement of 3.5 in AP. Both methods use an input size of $256\times192$ and no Online Hard Keypoints Mining(OHKM) involved.

CPN~\cite{chen2018cascaded} and our baseline use the same backbone of ResNet-50. When OHKM is not used, our baseline outperforms CPN~\cite{chen2018cascaded} by 1.8 AP for input size $256\times192$, and 1.6 AP for input size $384\times288$. When OHKM is used in CPN~\cite{chen2018cascaded}, our baseline is better by 0.6 AP for both input sizes.

Note that the results of Hourglass~\cite{newell2016stacked} and CPN~\cite{chen2018cascaded} are cited from~\cite{chen2018cascaded} and not implemented by us. Therefore, the performance difference could come from implementation difference. Nevertheless, we believe it is safe to conclude that our baseline has comparable results but is simpler.

\setlength{\tabcolsep}{2pt}
\begin{table}[t]
\begin{center}
\caption{Comparisons on COCO test-dev dataset. \textbf{Top}: methods in the literature, trained only on COCO training dataset. \textbf{Middle}: results submitted to COCO test-dev leaderboard~\cite{cocoleaderboard}, which have either extra training data (*) or models ensamled ($^+$). \textbf{Bottom}: our single model results, trained only on COCO training dataset.}
\label{table:coco_result}
\begin{tabular}{lllllllll}
\hline\noalign{\smallskip}
Method &Backbone& Input Size&$AP$ & $AP_{50}$&$AP_{75}$&$AP_{m}$&$AP_{l}$&$AR$\\
\noalign{\smallskip}
\hline
CMU-Pose~\cite{cao2017realtime} &-& -&$61.8$ & $84.9$&$67.5$&$57.1$&$68.2$&$66.5$\\
Mask-RCNN~\cite{he2017mask} & ResNet-50-FPN&-&$63.1$ & $87.3$&$68.7$&$57.8$&$71.4$&-\\
G-RMI~\cite{papandreou2017towards} & ResNet-101&$353\times257$&$64.9$ & $85.5$&$71.3$&$62.3$&$70.0$&$69.7$\\
CPN~\cite{chen2018cascaded} & ResNet-Inception&$384\times288$&$72.1$ & $91.4$&$80.0$&$68.7$&$77.2$&$78.5$\\
\hline
FAIR*~\cite{cocoleaderboard} &ResNeXt-101-FPN&- &$69.2$ & $90.4$&$77.0$&$64.9$&$76.3$&$75.2$\\
G-RMI*~\cite{cocoleaderboard} &ResNet-152&$353\times257$& $71.0$ & $87.9$&$77.7$&$69.0$&$75.2$&$75.8$\\
oks*~\cite{cocoleaderboard} &-&- &$72.0$ & $90.3$&$79.7$&$67.6$&$78.4$&$77.1$\\
bangbangren*$^+$~\cite{cocoleaderboard} &ResNet-101 &-&$72.8$ & $89.4$&$79.6$&$68.6$&$\textbf{80.0}$&$78.7$\\
CPN$^+$~\cite{chen2018cascaded,cocoleaderboard} & ResNet-Inception&$384\times288$&$73.0$ & $\textbf{91.7}$&$80.9$&$69.5$&$78.1$&$\textbf{79.0}$\\
\hline
Ours & ResNet-152&$384\times288$&$\textbf{73.7}$ & $\textbf{91.9}$&$\textbf{81.1}$&$\textbf{70.3}$&$\textbf{80.0}$&$\textbf{79.0}$\\
\hline
\end{tabular}
\end{center}
\end{table}
\setlength{\tabcolsep}{1.4pt}

\subsubsection{Comparisons on COCO test-dev dataset} 
Table~\ref{table:coco_result} summarizes the results of other state-of-the-art methods in the literature on COCO Keypoint Leaderboard~\cite{cocoleaderboard} and COCO \emph{test-dev} dataset. For our baseline here, a human detector with \emph{person detection} AP of 60.9 on COCO \emph{std-dev} split dataset is used. For reference, CPN~\cite{chen2018cascaded} use a human detector with \emph{person detection} AP of 62.9 on COCO \emph{minival} split dataset.

Compared with CMU-Pose~\cite{cao2017realtime}, which is a bottom-up approach for multi-person pose estimation, our method is significantly better. Both G-RMI~\cite{papandreou2017towards} and CPN~\cite{chen2018cascaded} have a similar top-down pipeline with ours. G-RMI also uses ResNet as backbone, as ours. Using the same backbone Resnet-101, our method outperforms G-RMI for both small ($256\times192$) and large input size ($384\times288$). CPN uses a stronger backbone of ResNet-Inception~\cite{szegedy2017inception}. As evidence, the top-1 error rate on ImageNet validation set of Resnet-Inception and ResNet-152 are $18.7\%$ and $21.4\%$ respectively~\cite{szegedy2017inception}. Yet, for the same input size $384\times288$, our result 73.7 outperforms both CPN's single model and their ensembled model, which have 72.1 and 73.0 respectively.

\subsection{Pose Estimation and Tracking on PoseTrack}
PoseTrack~\cite{andriluka2018posetrack} dataset is a large-scale benchmark for multi-person pose estimation and tracking in videos. It requires not only pose estimation in single frames, but also temporal tracking across frames. It contains 514 videos including 66,374 frames in total, split into 300, 50 and 208 videos for training, validation and test set respectively. For training videos, 30 frames from the center are annotated. For validation and test videos, besides 30 frames from the center, every fourth frame is also annotated for evaluating long range articulated tracking. The annotations include 15 body keypoints location, a unique person id and a head bounding box for each person instance.

The dataset has three tasks. Task 1 evaluates single-frame pose estimation using mean average precision (mAP) metric as is done in~\cite{pishchulin2016deepcut}. Task 2 also evaluates pose estimation but allows usage of temporal information across frames. Task 3 evaluates tracking using multi-object tracking metrics~\cite{bernardin2008evaluating}. As our tracking baseline uses temporal information, we report results on Task 2 and 3. Note that our pose estimation baseline also performs best on Task 1 but is not reported here for simplicity.

\subsubsection{Training} Our pose estimation model is fine-tuned from those pre-trained on COCO in Section~\ref{sec:exp_coco}. As only key points are annotated, we obtain the ground truth box of a person instance by extending the bounding box of its all key points by 15\% in length (7.5\% on both sides). The same data augmentation as in Section~\ref{sec:exp_coco} is used. During training, the base learning rate is 1e-4. It drops to 1e-5 at 10 epochs and 1e-6 at 15 epochs. There are 20 epochs in total. Other hyper parameters are the same as in Section~\ref{sec:exp_coco}.

\subsubsection{Testing} 
\begin{figure*}[!t]
	\centering
	\includegraphics[width=1.0\textwidth]{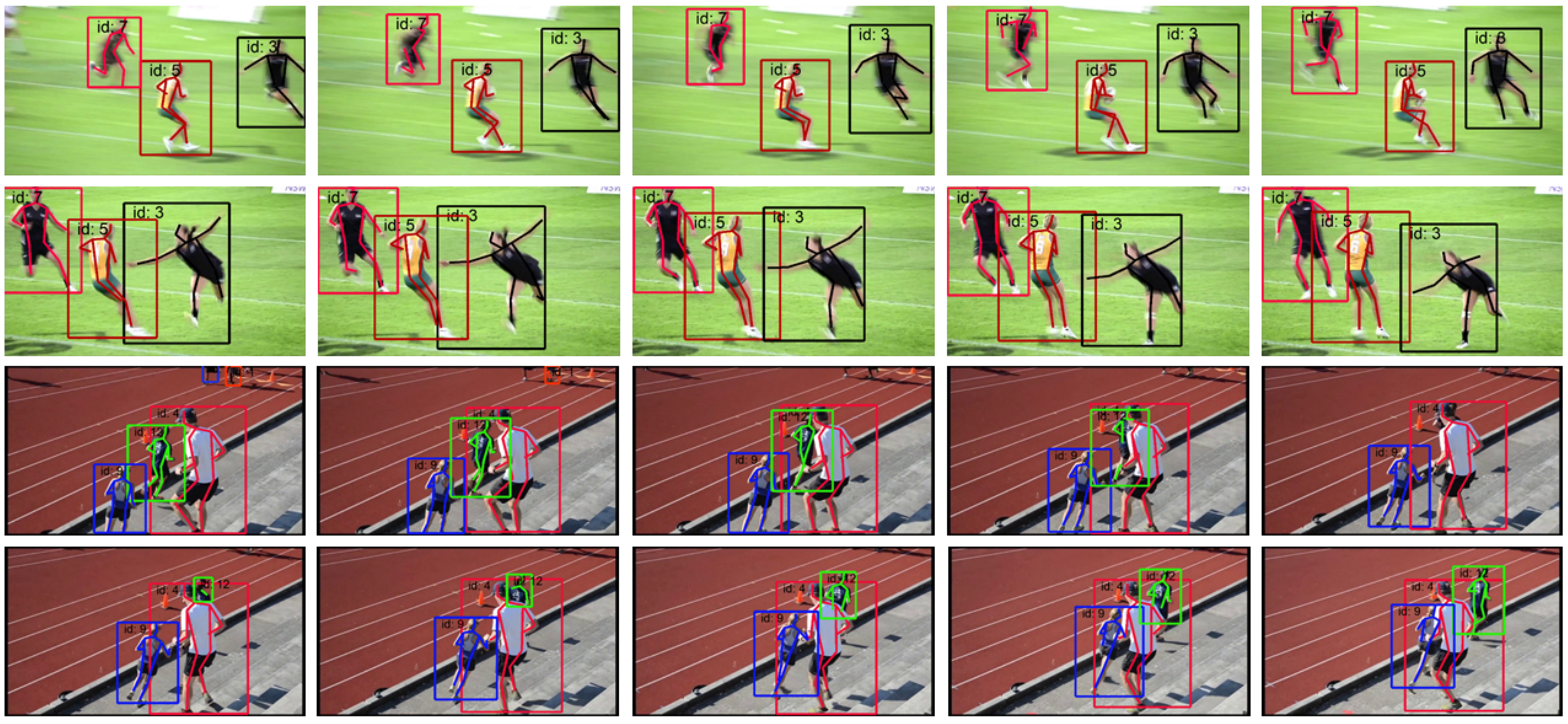}
	\caption{Some sample results on PoseTrack Challenge test set.}
	\label{fig:results_pose}
\end{figure*}
    
Our flow based tracking baseline is closely related to the human detector's performance, as the propagated boxes could affect boxes from a detector. To investigate its effect, we experiment with two off-the-shelf detectors, a faster but less accurate R-FCN~\cite{dai16rfcn} and a slower but more accurate FPN-DCN~\cite{dai17dcn}. Both use ResNet-101 backbone and are obtained from public implementation~\cite{dcn}. No additional fine tuning of detectors on PoseTrack dataset is performed.

Similar as in~\cite{girdhar2018detect}, we first drop low-confidence detections, which tends to decrease the mAP metric but increase the MOTA tracking metric. Also, since the tracking metric MOT penalizes false positives equally regardless of the scores, we drop low confidence joints first to generate the result as in~\cite{girdhar2018detect}. We choose the boxes and joints drop threshold in a data-driven manner on validation set, 0.5 and 0.4 respectively.

For optical flow estimation, the fastest model FlowNet2S in FlowNet family~\cite{ilg2017flownet} is used, as provided on~\cite{flownet2-pytorch}. We use the PoseTrack evaluation toolkit for results on validation set and report final results on test set from the evaluation server.
Fig.~\ref{fig:results_pose} illustrates some results of our approach on PoseTrack test dataset.

Our main ablation study is performed on ResNet-50 with input size $256\times192$, which is already strong when compared with state-of-the-art. Our best result is on ResNet-152 with input size $384\times288$. 

\begin{table}[t]
\centering
\caption{Ablation study on PoseTrack Challenge validation dataset. \textbf{Top}: Results of ResNet-50 backbone using R-FCN detector. \textbf{Middle}: Results of ResNet-50 backbone using FPN-DCN detector. \textbf{Bottom}: Results of ResNet-152 backbone using FPN-DCN detector.}
\label{table:posetrack_val_ablation_result}
\begin{tabular}{@{}lllclcc@{}}
\toprule
Method & Backbone & Detector & \tabincell{c}{With Joint \\Propagation} &\tabincell{c}{Similarity \\ Metric} & \tabincell{c}{mAP \\ Total} & \tabincell{c}{MOTA \\Total}\\ \midrule
\textbf{$a_1$} & ResNet-50 & R-FCN & \xmark & $S_{Bbox}$  & 66.0 & 57.6\\
\textbf{$a_2$} & ResNet-50 & R-FCN & \xmark & $S_{Pose}$  & 66.0 & 57.7\\
\textbf{$a_3$} & ResNet-50 & R-FCN & \cmark & $S_{Bbox}$  & 70.3 & 61.4\\
\textbf{$a_4$} & ResNet-50 & R-FCN & \cmark & $S_{Pose}$  & 70.3 & 61.8\\
\textbf{$a_5$} & ResNet-50 & R-FCN & \cmark & $S_{Flow}$  & 70.3 & 61.8\\
\textbf{$a_6$} & ResNet-50 & R-FCN & \cmark & $S_{Multi-Flow}$ & \textbf{70.3} & \textbf{62.2} \\ 
\midrule
\textbf{$b_1$} & ResNet-50 & FPN-DCN &\xmark & $S_{Bbox}$ &  69.3 & 59.8\\
\textbf{$b_2$} & ResNet-50 & FPN-DCN &\xmark & $S_{Pose}$ &  69.3 & 59.7\\
\textbf{$b_3$} & ResNet-50 & FPN-DCN &\cmark & $S_{Bbox}$ &  72.4 & 62.1\\
\textbf{$b_4$} & ResNet-50 & FPN-DCN &\cmark & $S_{Pose}$ &  72.4 & 61.8\\
\textbf{$b_5$} & ResNet-50 & FPN-DCN &\cmark & $S_{Flow}$ &  72.4 & 62.4\\
\textbf{$b_6$} & ResNet-50 & FPN-DCN &\cmark & $S_{Multi-Flow}$ &  \textbf{72.4} & \textbf{62.9} \\ 
\midrule
\textbf{$c_1$} & ResNet-152 & FPN-DCN & \xmark & $S_{Bbox}$ & 72.9 & 62.0\\
\textbf{$c_2$} & ResNet-152 & FPN-DCN & \xmark & $S_{Pose}$ & 72.9 & 61.9\\
\textbf{$c_3$} & ResNet-152 & FPN-DCN & \cmark & $S_{Bbox}$ & 76.7 & 64.8\\
\textbf{$c_4$} & ResNet-152 & FPN-DCN & \cmark & $S_{Pose}$ & 76.7 & 64.9\\
\textbf{$c_5$} & ResNet-152 & FPN-DCN & \cmark & $S_{Flow}$ & 76.7 & 65.1\\
\textbf{$c_6$} & ResNet-152 & FPN-DCN & \cmark & $S_{Multi-Flow}$ & \textbf{76.7} & \textbf{65.4}\\ \bottomrule

\end{tabular}
\end{table}

\setlength{\tabcolsep}{2.4pt}
\begin{table}[t]
\begin{center}
\caption{Multi-person Pose Estimation Performance on PoseTrack Challenge dataset. ``*'' means models trained on train+validation set. \textbf{Top}: Results on PoseTrack validation set. \textbf{Bottom}: Results on PoseTrack test set}
\label{table:posetrack_pose_result}
\begin{tabular}{lllllllllg}
\hline\noalign{\smallskip}
Method & Dataset&\tabincell{c}{Head \\ mAP}&\tabincell{c}{Sho. \\ mAP}&\tabincell{c}{Elb. \\ mAP}&\tabincell{c}{Wri. \\ mAP}&\tabincell{c}{Hip \\ mAP}&\tabincell{c}{Knee \\ mAP}&\tabincell{c}{Ank. \\ mAP}&\tabincell{c}{Total \\ mAP}\\
\noalign{\smallskip}
\hline
\noalign{\smallskip}
Girdhar et al.~\cite{girdhar2018detect} & val& 67.5&70.2&62.0&51.7&60.7&58.7&49.8&60.6\\
Xiu et al.~\cite{xiu2018pose} & val& 66.7&73.3&68.3&61.1&67.5&67.0&61.3&66.5\\
Ours:ResNet-50 & val& 79.1 & 80.5 & 75.5 & 66.0 & 70.8 & 70.0 & 61.7 & 72.4 \\
Ours:ResNet-152 & val& 81.7 & 83.4 & 80.0 & 72.4 & 75.3 & 74.8 & 67.1 & \textbf{76.7} \\
\hline
Girdhar et al.*~\cite{girdhar2018detect} & test& -&-&-&-&-&-&-&59.6\\
Xiu et al.~\cite{xiu2018pose} & test& 64.9&67.5&65.0&59.0&62.5&62.8&57.9&63.0\\
Ours:ResNet-50 & test & 76.4 & 77.2 & 72.2 & 65.1 & 68.5 & 66.9 & 60.3 & 70.0 \\
Ours:ResNet-152 & test& 79.5 & 79.7 & 76.4 & 70.7 & 71.6 & 71.3 & 64.9 & \textbf{73.9} \\
\hline
\end{tabular}
\end{center}
\end{table}
\setlength{\tabcolsep}{1.4pt}

\subsubsection{Effect of Joint Propagation} 
Table~\ref{table:posetrack_val_ablation_result} shows that using boxes from joint propagation introduces improvement on both mAP and MOTA metrics using different backbones and detectors. With R-FCN detector, using boxes from joint propagation (method \textbf{$a_3$} vs. \textbf{$a_1$}) introduces improvement of 4.3 \% mAP and 3.8 \% MOTA. With the better FPN-DCN detector, using boxes from joint propagation (method \textbf{$b_3$} vs. \textbf{$b_1$}) introduces improvement of 3.1 \%mAP and 2.3 \% MOTA. With ResNet-152 as backbone (method $c_3$ vs. $c_1$), improvement is 3.8 \% mAP and 2.8 \% MOTA. Note that such improvement does not only come from more boxes. As noted in~\cite{girdhar2018detect}, simply keeping more boxes of a detector, e.g., by using a smaller threshold, would lead to an improvement in mAP, but a drop in MOTA since more false positives would be introduced. The joint propagation improves both mAP and MOTA metrics, indicating that it finds more persons that are missed by the detector, possibly due to motion blur or occlusion in video frames.

Another interesting observation is that the less accurate R-FCN detector benefits more from joint propagation. For example, the gap between using FPN-DCN and R-FCN detector in ResNet-50 is decreased from 3.3\% mAP and 2.2\% MOTA (from $a_1$ to $b_1$) to 2.1\% mAP and 0.4\% MOTA (from $a_3$ to $b_3$). Also, method $a_3$ outperforms method $b_1$ by 1.0\% mAP and 1.6\% MOTA, indicating that a weak detector R-FCN combined with joint propagation could perform better than a strong detector FPN-DCN along. While, the former is more efficient as joint propagation is fast.

\subsubsection{Effect of Flow-based Pose Similarity}
Flow-based pose similarity is shown working better when compared with bounding box similarity and pose similarity in Table~\ref{table:posetrack_val_ablation_result}. For example, flow-based similarity using multi frames (method $b_6$) and single frame (method $b_5$) outperforms bounding box similarity (method $b_3$) by 0.8\% MOTA and 0.3\% MOTA. 

Note that flow-based pose similarity is better than bounding box similarity when person moves fast and their boxes do not overlap. Method $b_6$ with flow-based pose similarity considers multi frames and have an 0.5\% MOTA improvement when compared to method $b_5$, which considers only one previous frame. This improvement comes from the case when people are lost shortly due to occlusion and appear again.

\subsubsection{Comparison with State-of-the-Art}

\setlength{\tabcolsep}{2pt}
\begin{table}[t]
\tiny
\begin{center}
\caption{Multi-person Pose Tracking Performance on PoseTrack Challenge dataset.``*'' means models trained on train+validation set. \textbf{Top}: Results on PoseTrack validation set. \textbf{Bottom}: Results on PoseTrack test set}
\label{table:posetrack_tracking_result}
\begin{tabular}{lllllllllglll}
\hline\noalign{\smallskip}
Method & Dataset&\tabincell{c}{MOTA \\ Head}&\tabincell{c}{MOTA \\ Sho.}&\tabincell{c}{MOTA \\ Elb.}&\tabincell{c}{MOTA \\ Wri.}&\tabincell{c}{MOTA \\ Hip}&\tabincell{c}{MOTA \\ Knee}&\tabincell{c}{MOTA \\ Ank.}&\tabincell{c}{MOTA \\ Total}&\tabincell{c}{MOTP \\ Total}&\tabincell{c}{Prec \\ Total}&\tabincell{c}{Rec\\Total}\\
\noalign{\smallskip}
\hline
\noalign{\smallskip}
Girdhar et al.~\cite{girdhar2018detect} & val& 61.7&65.5&57.3&45.7&54.3&53.1&45.7&55.2&61.5&66.4&88.1\\
Xiu et al.~\cite{xiu2018pose} & val& 59.8&67.0&59.8&51.6&60.0&58.4&50.5&58.3&67.8&70.3&87.0\\
Ours:ResNet-50 & val& 72.1 & 74.0 & 61.2 & 53.4 & 62.4 & 61.6 & 50.7 & 62.9 & 84.5 & 86.3 & 76.0 \\
Ours:ResNet-152 & val& 73.9 & 75.9 & 63.7 & 56.1 & 65.5 & 65.1 & 53.5 & \textbf{65.4} & 85.4 & 85.5 & 80.3 \\
\noalign{\smallskip}
\hline
\noalign{\smallskip}
Girdhar et al.*~\cite{girdhar2018detect} & test& -&-&-&-&-&-&-&51.8&-&-&-\\
Xiu et al.~\cite{xiu2018pose} & test& 52.0&57.4& 52.8 &46.6&51.0&51.2&45.3&51.0&16.9&71.2&78.9\\
Ours:ResNet-50 & test& 65.9 & 67.0 & 51.5 & 48.0 & 56.2 & 54.6 & 46.9 & 56.4 & 45.5 & 81.0 & 75.7 \\
Ours:ResNet-152 & test& 67.1 & 68.4 & 52.2 & 48.9 & 56.1 & 56.6 & 48.8 & \textbf{57.6} & 62.6 & 79.4 & 79.9 \\
\noalign{\smallskip}
\hline
\end{tabular}
\end{center}
\end{table}

\begin{table}[t]
\setlength{\tabcolsep}{2pt}
\begin{center}
\caption{Results of Mulit-Person Pose Tracking on PoseTrack Challenge Leaderboard.``*'' means models trained on train+validation set.}
\label{table:posetrack_leaderboard}

\begin{tabular}{llcc}
\hline\noalign{\smallskip}
Entry &Additional Training Dataset&mAP&MOTA\\
\noalign{\smallskip}
\hline
ProTracker~\cite{girdhar2018detect}&COCO&59.6&51.8\\
PoseFlow~\cite{posetrackleaderboard}&COCO+MPII-Pose&63.0&51.0\\
MVIG~\cite{posetrackleaderboard}&COCO+MPII-Pose&63.2&50.7\\
BUTD2~\cite{jin2017towards}&COCO&59.2&50.6\\
SOPT-PT~\cite{posetrackleaderboard}&COCO+MPII-Pose&58.2&42.0\\
ML-LAB~\cite{zhu2017multi}&COCO+MPII-Pose&70.3&41.8\\
\hline
Ours:ResNet152*&COCO& \textbf{74.6} & \textbf{57.8} \\
\hline
\end{tabular}
\end{center}
\end{table}

We report our results on both Task 2 and Task 3 on PoseTrack dataset. As verified in Table~\ref{table:posetrack_val_ablation_result}, method $b_6$ and $c_6$ are the best settings and used here. Backbones are ResNet-50 and ResNet-152, respectively. The detector is FPN-DCN~\cite{dai17dcn}.

Table~\ref{table:posetrack_pose_result} reports the results on pose estimation (Task 2). Our small model (ResNet-50) outperforms the other methods already by a large margin. Our larger model (ResNet-152) further improves the state-of-the-art. On validation set it has an absolute 16.1\% improvement in mAP over~\cite{girdhar2018detect}, which is the winner of ICCV'17 PoseTrack Challenge, and also has an 10.2\% improvement over a recent work~\cite{xiu2018pose}, which is the previous best.

Table~\ref{table:posetrack_tracking_result} reports the results on pose tracking (Task 3). Compared with~\cite{girdhar2018detect} on validation and test dataset, our larger model (ResNet-152) has an 10.2 and 5.8 improvement in MOTA over its 55.2 and 51.8 respectively. Compared with the recent work~\cite{xiu2018pose}, our best model (ResNet-152) has 7.1\% and 6.6\% improvement on validation and test dataset respectively. Note that our smaller model (ResNet-50) also outperform the other methods~\cite{girdhar2018detect,xiu2018pose}. 

Table~\ref{table:posetrack_leaderboard} summarizes the results on PoseTrack's leaderboard. Our baseline outperforms all public entries by a large margin. Note that all methods differ significantly and this comparison is only on system level.

\section{Conclusions}

We present simple and strong baselines for pose estimation and tracking. They achieve state-of-the-art results on challenging benchmarks. They are validated via comprehensive ablation studies. We hope such baselines would benefit the field by easing the idea development and evaluation.

\bibliographystyle{splncs04}
\bibliography{egbib}
\end{document}